\documentclass[letterpaper, 10 pt, conference]{ieeeconf}  
\usepackage[colorlinks = true,
            linkcolor = blue,
            urlcolor  = blue,
            citecolor = blue,
            anchorcolor = blue]{hyperref}
\usepackage{graphicx}
\usepackage{subcaption}
\usepackage{multicol, latexsym, amsmath, amssymb}
\usepackage{multirow}
 \usepackage{booktabs}
\usepackage{tabularx}
\usepackage[noadjust]{cite}
\usepackage{etoolbox,lipsum}
\usepackage[format=plain]{caption}
\usepackage{adjustbox}
\let\labelindent\relax
\usepackage{enumitem}
\usepackage{wrapfig}
\usepackage{caption}
\usepackage{algorithm}
\usepackage{algpseudocode}
\usepackage{threeparttable}

\usepackage[most]{tcolorbox}
\usepackage{xcolor}
\usepackage{lipsum}

\IEEEoverridecommandlockouts                              

\title{\LARGE \bf
SmallPlan: Leverage Small Language Models for
Sequential Path Planning with LLM-Guided Distillation and Reinforcement Learning from Simulation Feedback
}

\author{Quang P. M. Pham${}^{1, +}$, Khoi T. N. Nguyen${}^{2,+}$, \\ Nhi H. Doan${}^{1}$,
Cuong A. Pham${}^{1}$, Qinbo Sun${}^{1}$, Weimin Qi${}^{1}$, Kentaro Inui${}^{1}$, Dezhen Song${}^{1, *}$%
\thanks{$^{1}$Department of Robotics, MBZUAI, UAE
    {\tt\small \{quang.pham, nhi.doan, cuong.pham, qinbo.sun, weimin.qi dezhen.song\}@mbzuai.ac.ae}}%
\thanks{$^{2}$
    {\tt\small nguyentietnguyenkhoi@gmail.com}}%
\thanks{$^{+}$Equal contribution}%
\thanks{$^*$Corresponding author}
}

\begin{document}

\maketitle

\begin{abstract}
    Efficient path planning in robotics, particularly within large-scale, complex environments, remains a significant hurdle. While Large Language Models (LLMs) offer strong reasoning capabilities, their high computational cost and limited adaptability hinder real-time deployment on edge devices. We present \textbf{SmallPlan} - a novel framework leveraging LLMs as teacher models to train lightweight Small Language Models (SLMs) for high-level path planning tasks. In SmallPlan, the SLMs provide optimal action sequences to navigate across scene graphs that compactly represent full-scaled 3D scenes. The SLMs are trained in a simulation-powered, interleaved manner with LLM-guided supervised fine-tuning (SFT) and reinforcement learning (RL). This strategy not only enables SLMs to successfully complete navigation tasks but also makes them aware of important factors like distance travel, providing more efficient path planning. Through experiments, we demonstrate that the fine-tuned SLMs perform competitively with larger models like GPT-4o on sequential path planning, without suffering from hallucination and overfitting. SmallPlan is resource-efficient, making it well-suited for edge-device deployment and advancing practical autonomous robotics. Our source code is available here: \hyperlink{https://github.com/quangpham2006/SmallPlan}{\texttt{https://github.com/quangpham2006/SmallPlan}}

\end{abstract}

\section{Introduction}
Recent advances in LLMs have played a key role in pushing the boundaries of robotic fields, particularly in scene understanding and path planning, due to their capability in semantic reasoning and problem-solving. \cite{pmlr-v205-ichter23a, 10235949, driess2023palmeembodiedmultimodallanguage} works on foundational models that utilize LLMs for common-sense reasoning and task planning. Leveraging these works, \cite{yang2024guidinglonghorizontaskmotion, sermanet2024robovqa} propose Vision-Language Models for long-horizon planning tasks, aiming to enable robots to perceive and traverse large-scale and complex environments. More recent studies \cite{rana2023sayplan, chen2024mapgptmapguidedpromptingadaptive} explore LLMs to generate long-horizon navigation plans from topological maps \cite{duggal2022topologically, chaplot2020neural} or graph-based scene representations (scene graph) \cite{pham2024esgnn, pham2024tesgnn, wu2021scenegraphfusionincremental3dscene, 10016416}. Compared to frame-by-frame reasoning with full-scale 3D representations, these approaches offer a lighter and compact alternative while retaining critical global semantic and spatial context.
\begin{figure}[ht]
    \centering
    \includegraphics[width=0.495\textwidth]{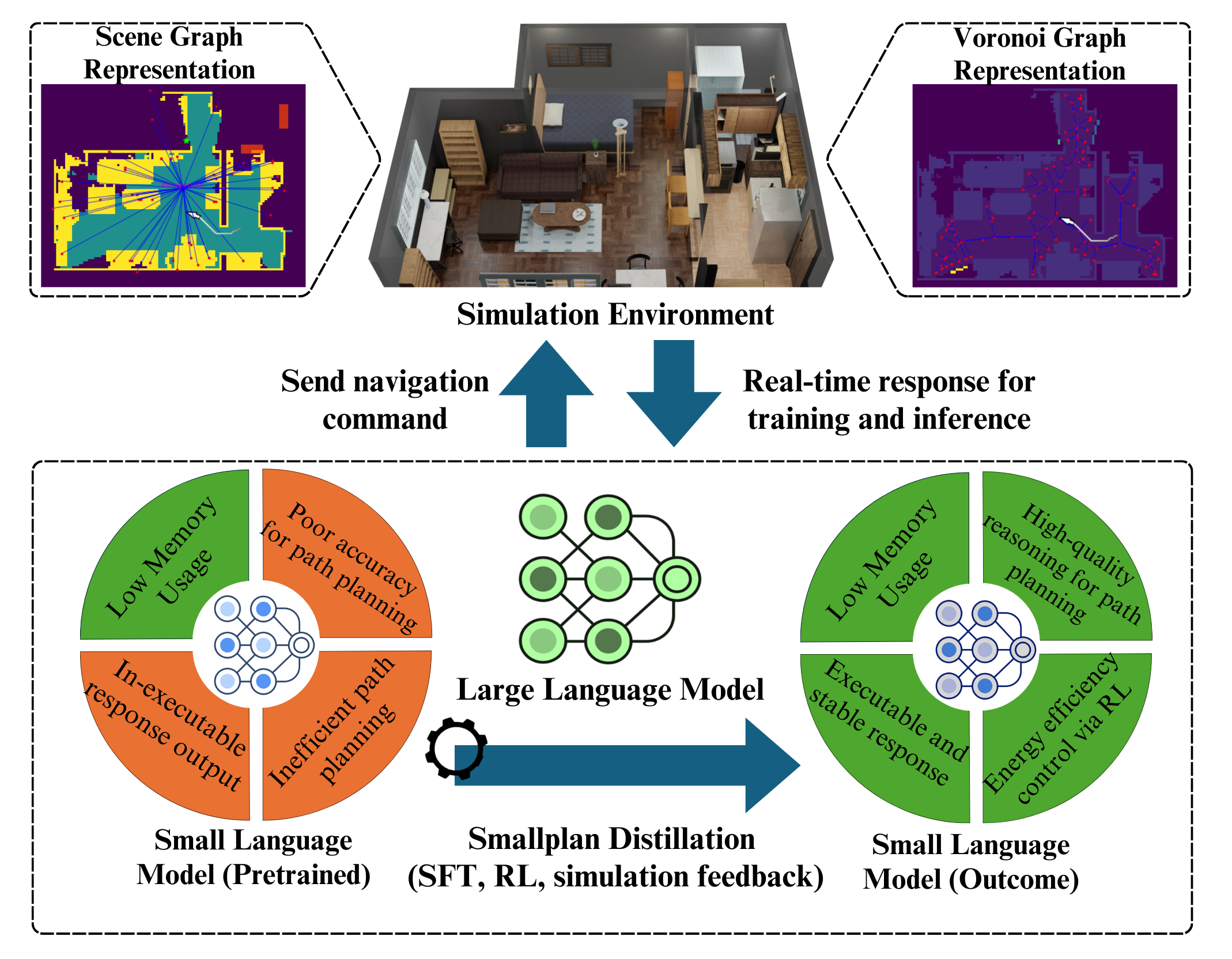}
    \caption{LLMs are powerful but often costly, SLMs are efficient but lack reasoning. We bridge the gap—training SLMs for sequential path planning leveraging LLMs, scene graph, SFT, and RL.}
    \label{fig:teaser}
    \vspace{-5mm}
\end{figure}

However, these approaches still rely either on heavy-weight LLMs, which are computationally expensive, or on LLM-as-a-Service (GPT-4o, GPT-4) \cite{rana2023sayplan, Honerkamp_2024}, which incur high API costs. Such setups are impractical for edge deployment and raise concerns about carbon emissions \cite{patterson2021carbon}. In addition, using off-the-shelf LLMs for direct inference makes it difficult to control important factors such as the distance or cost of the planning sequences they generate, which are crucial for managing energy consumption on edge devices.


Therefore, we propose \textbf{SmallPlan}, a distillation framework that uses LLMs as teacher models to train the SLMs for high-level path planning. Our framework enhances the outputs of SLMs by incorporating two key components. Firstly, the candidate SLM learns not only from the teacher LLM but also from real-time simulation feedback during training, achieving a balance between following a stronger model and adapting to surrounding environments. Secondly, we leverage RL and design a custom reward policy that incentivizes the SLM to control the travel distance and number of trials, making the inference less resource-consuming.

Our key contributions, demonstrated in Fig. \ref{fig:teaser}, include:
\begin{itemize}
    \item We enhance SLMs for sequential high-level path planning by combining LLM-guided distillation fine-tuning and RL from simulation feedback.
    
    \item We show that SLMs can achieve competitive performance with LLMs despite significantly lower resource requirements. SmallPlan enhances the SLMs' capability by learning from superior LLMs. Moreover, with RL and our reward policy, SmallPlan controls important factors to make the planning more efficient, enhancing real-world deployability.
    
    \item We study different training strategies that contribute to the outcome SLMs. We analyze the critical role of SFT distillation in stabilizing the SLMs' outputs, and how the SLMs adapt to different reward penalties in RL.
    
    
    
\end{itemize}

\section{Related Works}
\label{sec:related works}

\textbf{Scene Understanding and Representation.} A primary challenge in path planning is enabling robots to perceive and traverse large-scale and complex environments. VLMs have become a prevalent solution by leveraging foundation model reasoning for long-horizon planning tasks \cite{yang2024guidinglonghorizontaskmotion, sermanet2024robovqa}. However, these approaches often operate on a per-frame basis, lacking global scene understanding. Such localized processing limits their performance on large-scale navigation tasks, as individual frames provide only partial and fragmented views of the surroundings. Therefore, recent efforts utilized topological maps \cite{duggal2022topologically, chaplot2020neural} and scene graphs \cite{pham2024esgnn, pham2024tesgnn, wu2021scenegraphfusionincremental3dscene, 10016416} to encode semantic and spatial relationships from raw sensor data (point clouds, videos, images). From these compact representations, LLMs can be applied for higher level tasks. MapGPT \cite{chen2024mapgptmapguidedpromptingadaptive} integrates topological maps, vision models, and LLM prompt management, to enhance adaptive path planning. SayPlan \cite{rana2023sayplan}, MoMaLLM \cite{Honerkamp_2024} employ scene graphs to generate long-horizon navigation plans with LLMs. Our approach follows this direction, using scene graph instead of full-scale 3D representation, due to its efficiency and effectiveness in representing semantic and spatial relationships.

\textbf{LLMs for Planning and Reasoning Tasks.} Despite exhibiting impressive capability, LLMs sometimes struggle with reasoning tasks that are relatively simple for humans. To address this, \cite{wei2022chain, wang2022self, yao2022react, zhou2022least} experiment the effect of different prompt templates in LLMs' reasoning, in search of optimal prompting strategies. More recent research, on the other hand, highlights the adaptability of LLMs to human-defined rules \cite{zhu2023large}, suggesting promising applications in long-horizon planning tasks in robotics \cite{wang2024robogenunleashinginfinitedata, Honerkamp_2024, ahn2024autortembodiedfoundationmodels, huang2024rekepspatiotemporalreasoningrelational}. However, existing methods rely on pre-trained LLMs for direct reasoning without fine-tuning, leading to inefficient re-planning and inexecutable action sequences \cite{rana2023sayplan}. Additionally, they often require iterative planning cycles to produce executable plans, which is computationally intensive and cost inefficient for on-device inference. Our approach aims to replaces LLMs with lighter SLMs while retaining strong reasoning capability with knowledge distillation.

\textbf{Definition of SLMs.} Categorizations of SLMs varies across contexts \cite{wang2024comprehensivesurveysmalllanguage}. For example, \cite{liu2024mobilellm} defines SLMs as models with fewer than 1 billion parameters, while \cite{lee-etal-2024-small, tang2024smalllanguagemodelseffective, wang2024canslm} consider the term SLMs relative to their larger counterparts. For mobile devices, typically possessing around 6GB of memory, \cite{pmlr-v202-fu23d2023} suggests SLMs should have fewer than 10 billion parameters. NVIDIA \cite{belcak2025smalllanguagemodelsfuture} similarly considers models under 10 billion parameters as SLMs. In this work, we study the SLMs with at most 4 billion parameters and require less than 4GB of memory for inference.

\textbf{Distillation for SLMs.} Multiple studies \cite{lyu2024knowtuningknowledgeawarefinetuninglarge, wu2024divideorconquerdistillllm, ranaldi2024aligning, tian2024tinyllmlearningsmallstudent, mitra2023orca2teachingsmall} have explored techniques for transferring knowledge from large models (LLMs) to smaller ones (SLMs). One common strategy is feedback-driven fine-tuning \cite{guo2024directlanguagemodelalignment}. Other works \cite{agarwal2024onpolicy, yu2023languagerewardsroboticskill} suggest that feedback from mistakes and environment can improve fine-tuning in robotics task planning. \cite{fastnav} proposes a SFT distillation framework for turning point robot navigation. To further explore the capability of SLMs, we propose a multi-objective approach by additionally consider the SLMs' behaviors in simulation environment to reward the model with RL.

\textbf{SFT and RL for Fine-tuning\footnote{In our paper, "fine-tuning" and "training" are using interchangeably} Language Models.} 
Numerous works have investigated how SFT and RL enhance the reasoning capability of language models. An interesting study \cite{chu2025sftmemorizesrlgeneralizes} suggests that SFT tends to lead models to memorize the training data, whereas RL, especially when guided by outcome-based rewards, enables models to generalize across both rule-based textual and visual variants, thus enhancing the language model's outcomes. Notably, the study highlights that SFT remains essential for stabilizing the model's output format, thereby supporting reliable and effective RL training. \cite{dang2025reinforcementlearningreasoningsmall} specifically surveys what works and what does not for RL for reasoning in SLMs. Inspired by these findings, our work proposes a fine-tuning strategy that utilize SFT interleaved with reward-customized RL to stabilize and enhance the outcomes of the SLMs for path planning. 

\begin{figure*}[htbp]
    \centering
    \includegraphics[width=0.9\textwidth]{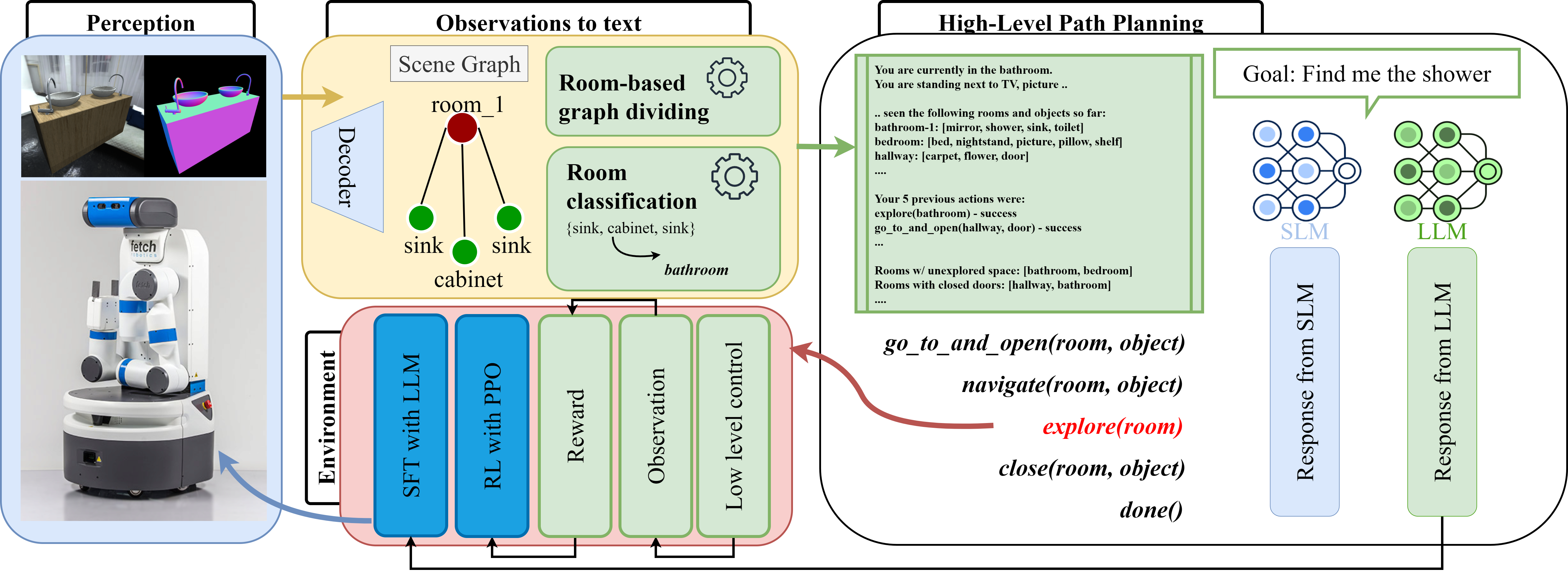}
    \caption{SmallPlan training framework. (Perception) A hierarchical scene graph constructed from robot observations, representing rooms and their associated objects. (Observations to Text) The scene graph then converted into text and feed to language models for path planning. (High-Level Path Planning) The LLM and SLM use the graph as context to propose the next actions. (Environment) The robot executes actions, receives new observations, and updates the SLM using rewards.}

    \label{fig:overall}
\end{figure*}

\section{Problem Statement and Scenario Set-up}
\label{sec:problem_statement}

We consider a scenario in which an embodied robotic agent $\mathcal{M}$ operates in a large, previously unexplored environment $\textbf{Env}$. The agent’s objective is to locate a target object $o_g \in \mathcal{O}$, where $\mathcal{O}$ denotes the complete set of objects within $\textbf{Env}$. To achieve this, $\mathcal{M}$ leverages a language model (SLM) to reason about $\textbf{Env}$ and generate sequential navigation actions that guide it toward $g$. Here, sequential navigation refers to a sequence of correlated actions drawn from an available action pool, where each chosen action builds upon the previous ones. This enables $\mathcal{M}$ to interact with both objects and the environment in a coherent manner to successfully complete the task. The following assumptions and constraints are imposed:

\textbf{Environment Scenario.} The environment $\textbf{Env}$ consists of two disjoint sets of objects: $\mathcal{O}_{\text{seen}}$, representing objects currently within the visible field of view of the robot $\mathcal{R}$, and $\mathcal{O}_{\text{unseen}}$, denoting objects outside of this view. These sets satisfy $\mathcal{O} = \mathcal{O}_{\text{seen}} \cup \mathcal{O}_{\text{unseen}}$ and $\mathcal{O}_{\text{seen}} \cap \mathcal{O}_{\text{unseen}} = \emptyset$. Transitions from $\mathcal{O}_{\text{unseen}}$ to $\mathcal{O}_{\text{seen}}$ occur through real-time perception using image or point cloud data. In the simulation environment, visibility is determined directly by the simulator engine, which updates the sets whenever an object enters the field of view of $\mathcal{M}$.

\textbf{Interactive Object Search.} Extending the setup in \cite{Honerkamp_2024}, our task builds on prior work \cite{chaplot2020learning, zhou2023esc, chen2023not} by requiring the robot $\mathcal{M}$ to interact with the environment $\textbf{Env}$—for example, opening doors or containers—to uncover occluded or hidden objects. This reflects realistic scenarios where the target object $g$ may be blocked by physical barriers such as cabinets or closed doors.

\label{sec:method}

\textbf{Action Space.} The agent operates over a discrete action space $\mathcal{A}$, which includes: \texttt{navigate(room\_name, object\_name)} to move toward a location; \texttt{go\_to\_and\_open(room\_name, object\_name)} for manipulating articulated objects; \texttt{close()} to shut opened objects; \texttt{explore(room\_name)} to search unexplored areas; and \texttt{done()} to terminate the task. In simulation, we assume perfect localization, while in real-world settings, SLAM systems such as ORB-SLAM \cite{Mur_Artal_2015} or Fast-LIO \cite{xu2021fastliofastrobustlidarinertial} can provide reliable pose estimation.

\textbf{Resource Constraints.} We assume $\mathcal{M}$ runs real-time on edge devices, with 6GB of memory. In our work, the SLMs are capped at 4 billion parameters with less than 4GB of memory for inference.

\section{Proposed Method}


Fig.~\ref{fig:overall} demonstrates SmallPlan's pipeline. At first, robot perception is fed into a model that contains a two-level hierarchical graph representing the scenes, with higher level represents a room, and lower level represents the objects located within that corresponding room. In simulation setup, information of the scenes and the outcomes of the proposed action sequences by the SLMs are extracted directly from the simulation engine. We apply chain-of-thought prompting~\cite{wei2022chain} similarly to~\cite{Honerkamp_2024} to generate responses for both the LLM and the SLM. The LLM and SLM responses, together with outcomes from the simulation environment, are utilized for the fine-tuning process with SFT and RL, with details provided in Section \ref{ssec: finetuning_pipeline}.


\subsection{From Perception to Language Model}
To provide effective text-based prior knowledge to language models, we transform robot observations into a text-based hierarchical scene graph. The process begins with constructing the Navigation Voronoi Graph $\mathcal{G_V} = (\mathcal{V}, \mathcal{E})$, where the Voronoi graph represents the medial-axis skeleton of free space, with nodes $\mathcal{V}$ and edges $\mathcal{E}$ capturing connectivity, following the approach introduced in Hydra~\cite{hughes2022hydra}. Inspired by \cite{Honerkamp_2024}, we then decompose $\mathcal{G_V}$ into multiple room-based Navigation Voronoi Graphs $\mathcal{G_V^R}$ by removing edges and nodes near doorways. These subgraphs are subsequently connected using the shortest path between a pair of nodes from each $\mathcal{G_V^R}$. Based on these $\mathcal{G_V^R}$, we transfer it to a two-level hierarchical scene graph $\mathcal{G_S}$, which captures two levels of abstraction, rooms and objects. Here $\mathcal{G_S} = (\mathcal{N, E})$, where $\mathcal{N} \in (\mathcal{N}_{room}, \mathcal{N}_{objects})$, in which the $\mathcal{N}_{objects}$ representing the objects discovered by robots $\mathcal{O}_{seen}$ and $\mathcal{N}_{room}$ is the room label $room\_name$, while $\mathcal{E}$ is the edge connecting $\mathcal{N}_{room}$ and $\mathcal{N}_{objects}$. We denote the environment state $s = \{\mathcal{G_V}, \mathcal{G_S}, \textbf{Env}\}$ includes both the Navigation Voronoi Graph and Scene Graph, and other environment information.

For each room-specific subgraph $\mathcal{G_V^R}$, we prompt the language model to classify the room type (e.g., living room, bathroom, kitchen, etc.), using the label other-room if the classification is uncertain. Next, we provide the language model with a list of rooms and their associated objects $\mathcal{O^R}$ can be seen from $\mathcal{G_V^R}$ $(\mathcal{O^R} \in \mathcal{N^R} \in \mathcal{G_S^R})$. Additionally, the model is given the previous action as context. When tasked with locating a specific object, the language model must respond with the next action $a \in \mathcal{A}$. To enhance reasoning, we apply a chain-of-thought approach, prompting the model to generate a structured response including analysis, reasoning, and a final command. For each action, the robot will execute the action, note that $\mathcal{G_V}$ is used for navigation purposes. 

\subsection{LLM-guided Distillation and RL with Simulation Feedback}
\label{ssec: finetuning_pipeline}



\subsubsection{Supervised Fine-tuning with teacher LLM} Assuming that the teacher LLM possesses superior reasoning and planning capabilities, we treat its response as the supervision label for the candidate SLM to follow. This is the distillation. We query both the teacher LLM and the candidate SLM for action predictions. We adopt the standard Cross-Entropy Loss as the \textbf{Supervised Loss} (\( \mathcal{L}_{S} \)) for aligning the outputs of the candidate SLM and the teacher LLM:
\begin{equation*}
    \mathcal{L}_{S} = -\sum_{i} y_i \log(\hat{y}_i),
\end{equation*}
where \( y_i \) represents the token from the LLM's response (used as the target label), and \( \hat{y}_i \) denotes the predicted probability assigned by the SLM for that token.

\subsubsection{Reinforcement Learning with Simulation Feedback} We reinforce the candidate SLM with the Proximal Policy Optimization (PPO) algorithm with a customized reward function, defined as:
$$
r_{\text{task}}(a_t, s_{t-1}, s_t) = 
\begin{cases} 
r_{\text{success}}, & \text{if task succeeds}, \\
r_{\text{action}}(a_t, s_{t-1}, s_t), & \text{otherwise.}
\end{cases}
$$
where \( a_t \) denotes the action taken at time step \( t \), $s_{t-1}$ is the environment state prior to the action and $s_t$ is the updated environment state after the action execution. A bonus reward \( r_{\text{success}} \) is granted if the task is successfully completed. The task is considered successful when the action selected by the SLM is \( a = \texttt{done()} \), and the resulting graph \( \mathcal{G}_{\mathcal{S}(t)} \) contains the goal object \( o_g \in (\mathcal{O}_{\text{seen}} = \mathcal{N}_{\text{objects}}) \subseteq \mathcal{G}_{\mathcal{S}(t)} \). Otherwise, rewards or penalties are assigned based on the outcomes of individual sub-actions during task execution. The action-specific reward \( r_{\text{action}} \):

\begin{align*}
    r_{\text{action}}(a_t, s_{t-1}, s_t) &=   r_{\text{action\_executable}}(a_t, s_{t-1}, s_t)
    \\ & + r_{\text{explore}}(a_t, s_{t-1}, s_t) 
    \\ & + r_{\text{efficiency}}(a_t, s_{t-1}, s_t)
    \\ & + r_\text{format}(a_t, s_{t-1}, s_t)
\end{align*}
    
Here, \( r_{\text{action\_executable}} \) provides a reward when the action is executable in the environment.  
\( r_{\text{explore}} \) encourages the robot to explore new areas, which is essential for locating the target object.  
\( r_{\text{efficiency}} \) penalizes excessive movement to promote efficient navigation and task execution.  
Finally, \( r_{\text{format}} \) imposes a penalty if the response generated by the SLM does not follow the required output format. Each component of the reward function is defined as:

\begin{itemize}
    \item Action Executable:
\begin{align*}
r_{\text{action\_executable}}(a_t, s_{t-1}, s_t) =
\begin{cases} 
    \lambda_{\text{executable}}, \ \text{if the action is} \\  \ \ \text{successfully executed,} \\
    -\lambda_{\text{executable}}, \ \text{otherwise.}
\end{cases}
\end{align*}
    
    The actions proposed by the SLM are not guaranteed to be executable by the robot \( \mathcal{R} \). For instance, the action \textit{go\_to\_and\_open(livingroom, cabinet)} is invalid if the \textit{cabinet} does not exist in the scene graph \( \mathcal{G_S}^{\text{livingroom}} \). This reward component encourages the model to reason based on the current observation \( \mathcal{G}_S \) and to produce feasible actions.

    \item Exploration:
    \[
    r_{\text{explore}}(a_t, s_{t-1}, s_t) = \lambda_{\text{explore}} \cdot \frac{\left\| \mathcal{V}_{\mathcal{G}_V(t)} - \mathcal{V}_{\mathcal{G}_V(t-1)} \right\|}{\eta_{\mathcal{G}_V}}
    \]
    Since the task involves locating a target object, exploration is encouraged. This reward is proportional to the number of newly discovered nodes in the visual scene graph \( \mathcal{G}_V \) after executing the action, where $\left\|\mathcal{V}_{\mathcal{G}_V(t)}\right\|$ is the number of nodes at time step $t$. \( \eta_{\mathcal{G}_V} \) is a normalization factor.

    \item Efficiency:
    \[
    r_{\text{efficiency}}(a_t, s_{t-1}, s_t) = \lambda_{\text{efficiency}} \cdot \frac{\left\| d_{\text{Env}(t)} - d_{\text{Env}(t-1)} \right\|}{\eta_d}
    \]
    To encourage energy efficiency and reduce unnecessary movement, this term penalizes the model based on the distance traveled. The total distance \(  d_{\text{Env}(t)} \) is retrieved from the simulation environment at time step $t$, and \( \eta_d \) is used for normalization.

    \item Format Adherence:
    \[
    r_{\text{format}}(a_t, s_{t-1}, s_t) = 
    \begin{cases} 
        0, \text{if the action is parsed} 
        \\ \text{ successfully,}
        \\
        -\lambda_{\text{format}}, \ \text{otherwise.}
    \end{cases}
    \]
    A penalty is applied if the SLM response cannot be parsed into a valid executable command.
\end{itemize}

where \(\lambda_{\text{action\_executable}}, \lambda_{\text{explore}}, \lambda_{\text{efficiency}}, \lambda_{\text{format}}\) are hyper-params that define the importance of each reward component.

\begin{algorithm}[ht]
\caption{SmallPlan Training Strategy}
\label{alg:training}
\begin{algorithmic}[1]
\Require $prompt\_template$, $LLM$, $SLM$, $num\_epochs$, $num\_epochs\_fewshot$, $tasks$
\vspace{0.4em}

\Statex \textbf{Stage 1: Few-shot SFT for Format Refinement}
\For{$epoch \gets 1$ \textbf{to} $num\_epochs\_fewshot$}
    \State $target\_ans \gets$ \textsc{LLM}.$\Call{answer}{prompt}$
    \State \textsc{SLM}.$\Call{sft\_train}{prompt, target\_ans}$
\EndFor
\vspace{0.4em}

\Statex \textbf{Stage 2: Interleaving RL--SFT Training}
\For{$epoch \gets 1$ \textbf{to} $num\_epochs$}
    \For{$task\_id$ \textbf{in} $tasks$} 
        \vspace{0.3em}
        \State $info \gets \Call{GetEnvInfo}{task\_id}$
        \State $prompt \gets \Call{AddTo}{prompt\_template, info}$
        \vspace{0.3em}
        \State $target\_ans \gets$ \textsc{LLM}.$\Call{answer}{prompt}$
        \State $slm\_ans \gets$ \textsc{SLM}.$\Call{answer}{prompt}$
        \State $todo\_actions \gets \Call{Parse}{slm\_ans}$
        \State $sim\_feedback \gets \Call{Execute}{todo\_actions}$
        \vspace{0.3em}
        \State \textsc{SLM}.$\Call{rl\_train}{slm\_ans, sim\_feedback}$
        \State \textsc{SLM}.$\Call{sft\_train}{prompt, target\_ans}$
    \EndFor
\EndFor

\end{algorithmic}
\end{algorithm}
\begin{table*}[htbp]
\centering
\begin{minipage}{0.65\linewidth} 
\caption{SmallPlan's SLMs, trained with RL-SFT strategy, across 7 test scenes with 25 runs per scene, randomized start poses, targets, and object distributions.}
\label{tab:evaluation_summary}
\centering
\scriptsize
\resizebox{\linewidth}{!}{%
\begin{tabular}{llcccc}
\toprule
\textbf{Baseline} & \textbf{Model}  & \textbf{SR \% $\uparrow$} & \textbf{SPL \% $\uparrow$} & \textbf{Dist. $\downarrow$} & \textbf{Retrials $\downarrow$} \\
\midrule
MoMaLLM \cite{Honerkamp_2024} & GPT-4o & \textbf{97.14} & \underline{65.92} & 18.01 & \textbf{0.39}\\
\addlinespace

Pretrained & Qwen-2.5 & 61.33 & 46.59 & 12.00 & 3.05 \\
Pretrained & Llama-3 & 53.68 & 38.02 & 9.84 & 4.16\\
Pretrained & Phi4-mini & 70.86 & 41.57 & 17.84 & 3.39\\
\addlinespace
Ours & Qwen-2.5 & 83.43 & 48.07 & 19.87 & 1.04\\
Ours & Llama-3 & 93.14 & 66.10 & 17.65 & 1.22\\
Ours & Phi4-mini & \underline{96.57} & \textbf{70.41} & \textbf{16.13} & \underline{0.49}\\
\bottomrule
\end{tabular}%
}
\end{minipage}
\vspace{-5mm}
\end{table*}

\subsubsection{SmallPlan Training Strategy} Algorithm \ref{alg:training} provides the pseudo-code for SmallPlan's training procedure, consisting of 2 stages. We employ Low-Rank Adaptation (LoRA) to accelerate the training process, which freezes the original model weights and focuses on modifying a smaller subset of weights to reduce computational and memory overhead.

In the \textbf{first stage}, the candidate SLM is supervised fine-tuned with a small set of response samples from the teacher LLM. Pretrained SLMs often fail to produce outputs in a consistent format, even with well-crafted instruction prompts, making it difficult to parse their responses into executable commands. This step thus helps the SLM learns the correct output format from the LLMs. This few-shot SFT strategy also fits with the LoRA setup as it helps the SLMs to highlight the important parameters for the target task. To speed up this stage, we collect the LLM responses and store them offline, and train SFT with these samples.

In the \textbf{second stage}, we train the SLM using an interleaved combination of RL and SFT. At runtime, we collect environment data including detected objects, their distances, and their relations from the scene graph and Voronoi graph, then parse this information into the prompt template to build a complete prompt as input for LLM and SLM to response. Here, SFT acts as a distillation mechanism for transferring reasoning behavior from the LLM, while RL - using our $r_{\text{action}}$ reward - adjusts the model’s planning behavior and controls task-related factors. This stage can also be configured to use only SFT, as distillation alone can sometimes be enough to achieve strong performance. In our experiments in Section \ref{sec:results}, we compare different training strategies, \textit{RL-SFT} and \textit{SFT only}, to evaluate their effectiveness. Our framework, however, strongly discourages removing the SFT and relying solely on RL, as this would prevent the SLM from effectively learning from the LLM.
\section{Experiments} \label{sec:results}

We study the following key points: (1) The effectiveness of SmallPlan in sequential path planning tasks; (2) The impact of different training strategy and trade-offs when using RL to control side factors; (3) the role of SFT during pre-training in enhancing subsequent RL training; and (4) How the trained SLMs generalize to out-of-distribution tasks.

\subsection{Experiment Setup}

\subsubsection{Model Selection} We investigate various SLMs, including \textbf{Qwen-2.5} (3.1B params) \cite{bai2023qwentechnicalreport}, \textbf{Llama-3} (3.2B) \cite{grattafiori2024llama3herdmodels}, and \textbf{Phi4-mini} (3.8B) \cite{abdin2024phi4technicalreport}. We used the dynamic 4-bit quantized versions provided by Unsloth AI. These models consume approximately 4GB of memory for inference. For smaller SLMs around 1B params, we discuss their performances in Section \ref{sec: limitations}. We compare SmallPlan's performance with the state-of-the-art MoMaLLM \cite{Honerkamp_2024}, which utilized GPT-4 as the language model for task planning.

\subsubsection{Simulation Setup and Dataset} We conduct experiments using the iGibson simulation platform \cite{shen2021igibson, li2022igibson} with the Fetch robot, adopting the setup from \cite{Honerkamp_2024}. The environment is configured based on the iGibson Challenge data split with 8 train scenes for development and 7 test scenes for evaluation. Each run within a scene randomizes the start poses, target objects, and object distributions, resulting in a diverse spectrum of task scenarios and action sequences. We evaluate our method over 25 run trials per test scene, yielding a total of 175 challenging and previously unseen task executions. 

\subsubsection{Out-of-domain Reasoning Analysis Setup} We conduct an out-of-domain experiment to assess the risk of overfitting in our fine-tuned SLMs. We use a text-based interactive game benchmark \cite{hudi2025textgameslearningselfplaytextbased}, designed to evaluate various logical reasoning skills of language models. Specifically, we compare the trained SLMs with their pretrained versions on 4 different games introduced in this benchmark. 

\subsubsection{Metrics} We evaluate SmallPlan using 4 metrics:
\vspace{1mm}

\noindent \textit{Task Success Rate (\textit{SR} \%)}: The percentage of successfully completed tasks out of all attempted trials. For out-of-domain analysis, this corresponds to the percentage of solved challenges out of all attempts.
\\
\noindent \textit{Success weighted by Path Length (SPL \%)}: The ratio of the shortest possible path length to the actual path taken, weighted by task success.
\\
\noindent \textit{Distance Traveled (Dist.)}: The total path length covered by the agent across all task executions.
\\
\noindent \textit{Average Retrials (Retrials)}: The average number of retries attempted by the agent per run per scene. 

\subsection{Results}

\begin{figure*}[ht]
    \centering
    \includegraphics[width=0.7\textwidth]{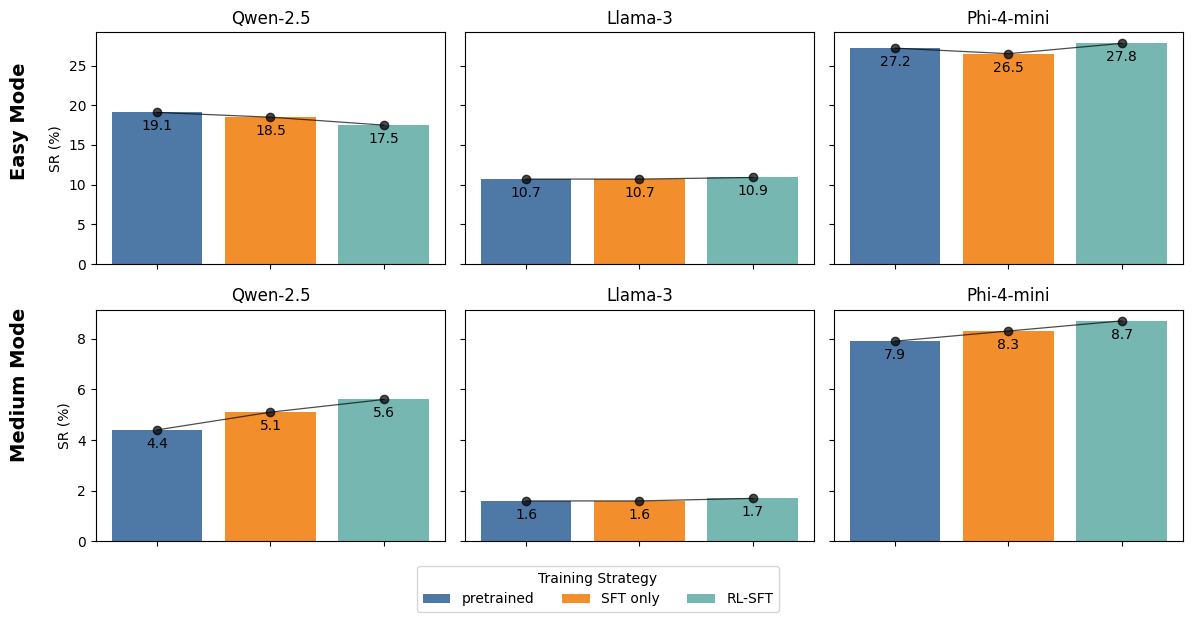}
    \caption{The mean \textit{SR (\%)} of the pre-trained and fine-tuned SLMs for the text-based games on Easy and Medium modes. There are 4 different games, with each game played 1,000 times with different sub-challenges provided by \cite{hudi2025textgameslearningselfplaytextbased}.}
    \label{fig:ood_plot}
    \vspace{-5mm}
\end{figure*}

\subsubsection{SmallPlan for Sequential Path Planning} \label{sssection: smallplan_sequential} While GPT-4o consistently achieves the highest \textit{SR}, our findings in Tables~\ref{tab:evaluation_summary} indicate that \textbf{significantly smaller models can also perform competitively on sequential path planning tasks}. Moreover, SmallPlan's SLMs are capable of optimizing side factors, as demonstrated by superior \textit{SPL} and \textit{Dist.} traveled metrics compared to GPT-4o. Note that, although some pretrained SLMs produce shorter \textit{Dist.}, their \textit{SR} is significantly lower due to command execution failures that prematurely reduce \textit{Dist.}, making this outcome not optimal. Some pretrained SLMs occasionally fail to follow the required output format in our prompts. For example, the pretrained Qwen-2.5 sometimes produces responses in Chinese rather than English. This issue is resolved after training with SmallPlan.

We ultimately show that fine-tuned SLMs perform reliably even when the starting poses, target objects, and object distributions are altered between runs. We also attribute the strong performance of Phi-4-mini to its pretraining on complex reasoning tasks \cite{microsoft2025phi4minitechnicalreportcompact}. Llama-3, despite not having the highest \textit{SR}, showing the largest \textit{SR} improvement over its pretrained version (Table \ref{tab:tradeoffs}), suggesting that this model can adapt and learn quite effectively. Qwen-2.5, while not matching the performance of ChatGPT or Phi4-mini, still show notable improvements over their pretrained versions.  

\begin{table}[h]
\caption{SmallPlan's SLMs on different training strategy.}
\label{tab:tradeoffs}
\centering
\footnotesize
\setlength{\tabcolsep}{2pt} 
\resizebox{0.95\linewidth}{!}{%
\begin{tabular}{llcccc}
\toprule
\textbf{Model} & \textbf{Strategy} & \textbf{SR \% $\uparrow$} & \textbf{SPL \% $\uparrow$} & \textbf{Dist. $\downarrow$} & \textbf{SRI \% $\uparrow$} \\
\midrule
\multirow{3}{*}{Qwen-2.5} & SFT only & 85.14 & 58.15 & 21.42 & 38.82\\
 & RL-SFT & 83.43 & 48.07 & 19.87 & 36.03\\
& RL-SFT $+$ & 78.86 & 59.31 & \underline{11.49} & 28.58\\
\addlinespace
\multirow{3}{*}{Llama-3} & SFT only & 88.57 & 64.36 & 13.54 & 64.99\\
 & RL-SFT & 93.14 & 66.10 & 17.65 & 73.50\\
  & RL-SFT $+$ & 74.30 & 57.83 & \underline{9.69} & 38.41\\
\addlinespace
\multirow{3}{*}{Phi4-mini} & SFT only & 91.43 & 58.40 & 18.26 & 29.02\\
 & RL-SFT & 96.57 & 70.41 & 16.13 & 36.28\\
& RL-SFT $+$ & 85.14 & 60.33 & \underline{13.52} & 20.15\\
\bottomrule
\end{tabular}%
}
\begin{tablenotes}
\footnotesize
 \item[*] ``RL-SFT"=balanced reward, ``RL-SFT $+$"=strong distance penalty. The \underline{underline} highlights the shortest \textit{Dist.} within same SLM architecture. \textit{SRI} denotes \% \textit{SR} increase compare to its pretrained version in Table \ref{tab:evaluation_summary}.
 \end{tablenotes}
\end{table}

\subsubsection{Trade-offs and Side Factor Control with RL} Table \ref{tab:tradeoffs} presents the performance of SLMs under different training strategies and reward penalties. We observe that applying RL does not always lead to higher \textit{SR}. This is reasonable, as SFT alone is sufficient to transfer the reasoning capability from LLMs to SLMs effectively. 

Notably, we highlight that \textbf{the candidate SLM adapt to different reward penalties from training}. For RL-SFT $^+$, we set the $\lambda_{\text{efficiency}}$ twice to that of RL-SFT $^*$ (0.6 vs. 0.3). This change result in a significant reduction in \textit{Dist.} traveled by the SLMs while retain better \textit{SR} compared to their pretrained versions, as shown in Table \ref{tab:tradeoffs}. There exist trade-off between \textit{SR} and \textit{Dist.}, as we observe that the SLMs tend to prematurely terminate tasks by calling \texttt{done()} instead of continuing to explore the environment. However, our approach paves the way for adaptation to different constraints and settings. While SFT trains the SLMs to reason by following the superior LLMs, it does not consider deployment constraints such as energy consumption. The SLMs trained with only SFT tend to re-prompt until reaching the maximum number of trials, whereas RL in our design allows early termination, reducing unnecessary computation. In real-time robotic applications on edge devices, each prompt incurs significant energy costs and action execution overhead, making efficient operation a key challenge. Thus, we consider a reasonable degree of the trade-off between \textit{SR.} and \textit{Dist.} acceptable. We conclude that \textbf{RL plays a crucial role in controlling side factors and enables more resource-efficient planning for deployment}.

\subsubsection{Few-shot SFT for Format Refinement.} In Algorithm \ref{alg:training}, we mentioned that SFT is applied before the main fine-tuning process for format refinement. In our experiment, removing this step results in unstable training dynamics and incoherent responses, as the scalar reward from RL alone is insufficient to guide learning. Therefore, we find that \textbf{training from the few-shot fine-tuned models is far more effective than from the pretrained ones, as the SFT stage establishes a strong prior for LoRA adaptation and output format adherence}.

\begin{figure*}[ht]
    \centering
    \includegraphics[width=0.99\textwidth]{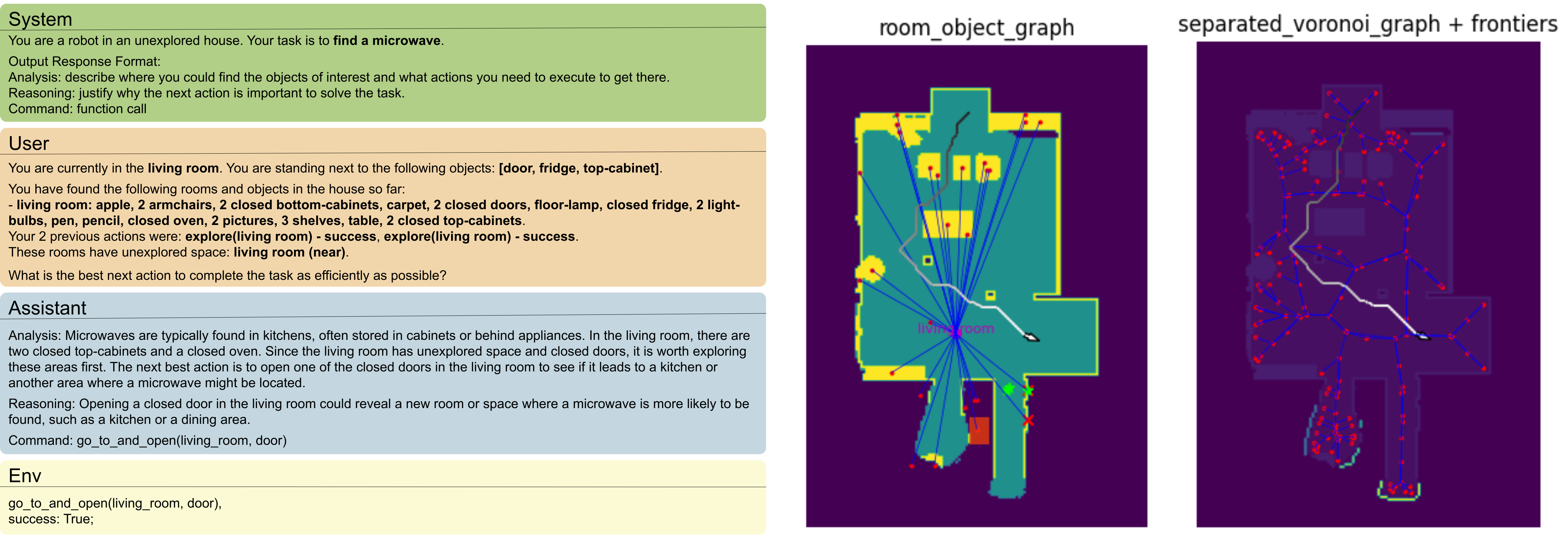}
    \caption{Inference response for task "find a microwave." The \textit{system} and \textit{user} blocks show the task-level and step-level prompts, \textit{assistant} shows the response by SmallPlan’s Phi-4-mini, \textit{env} shows environment feedback. The bold texts indicate the information feed to the prompt, including the task and info from the scene graph and Voronoi graph on the right side.}
    \label{fig:bevmap}
\vspace{-5mm}
\end{figure*}
\subsubsection{Out-of-domain Analysis} Fig. \ref{fig:ood_plot} evaluates the logical reasoning capabilities of the SLMs on out-of-domain tasks beyond path planning. These tasks involve different types of sequential decision-making, allowing us to assess general reasoning ability. Our trained SLMs perform on par with, and sometimes exceed, their pretrained versions, as reflected by the consistent \textit{SR} with only minor differences. This suggests that \textbf{distillation from a superior LLM can help retain the general reasoning ability of smaller models, even across diverse domains}. Given the complexity of cross-domain adaptation, we leave further investigation to future research. This experiment merely highlights that our trained SLMs maintain strong generalization and are not overfitting to path planning tasks. The goal here is not to compare different SLMs against each other on their success rates, but to compare each trained SLM with its pretrained version to show that our SLMs retain their general reasoning capability.

\subsubsection{Hyper-parameter Selection and Prompting} We try various hyper-parameter settings during training, including RL rewards and the number of training epochs. Based on our experience, selecting these settings is largely a trial-and-error process, as no standardized method can guarantee optimal results. We use $r_{\text{success}}=5$, $\lambda_{\text{action\_executable}}=0.3$, $\lambda_{\text{explore}}=0.1$, $\lambda_{\text{efficiency}}=0.3$, and $\lambda_{\text{format}}=0.1$. For penalization of \textit{Dist.} traveled, we increase the efficiency weight to $\lambda_{\text{efficiency}}=0.6$. Learning rate is fixed at $1 \times 10^{-5}$ for all training. Meanwhile, the number of epochs depends heavily on the SLM architectures, as they vary in learning capacity and convergence. For \textit{stage 1} (Algorithm \ref{alg:training}), we fine-tune with 3 epochs using 500 offline data samples for all SLMs. For \textit{stage 2}, we use 4 epochs for Phi-4-mini and Llama-3, and 6 epochs for Qwen-2.5. Simply increasing or decreasing the number of epochs does not consistently improve performance. We tested multiple settings and selected the best-performers. Our models perform best with 3 to 6 epochs. Further tuning efforts can help to find optimal hyper-parameters and further improve the SLMs' performance.

Fig. \ref{fig:bevmap} captures the real-time response from SmallPlan's Phi-4-mini for a specific task. In our prompt, we require the SLMs to provide the reason before proposing an action. This helps to improve the performance of the model. We observe that the SLM effectively follows the output format, as well as provide a concrete analysis and reasoning for its action from the provided context.


\section{Limitations and Discussions}
\label{sec: limitations}

\textbf{How small is enough?} In addition to our proposed SLMs, which are around 3-4B params, we also tested smaller models, including DeepSeek-R1 version 1.5B and Llama-3 version 1B. However, DeepSeek-R1 exhibited the same issue observed in Qwen-2.5 (Section~\ref{sssection: smallplan_sequential}) and producing poor results. Llama-3 (1B) achieved a \textit{SR} of 14.28\% for pretrained and 51.42\% for SmallPlan RL-SFT, increasing the \textit{SR} by 3.6 times. While trainable and more efficient, we believe these results are still insufficient for practical use. However, with the rapid effort made by NVIDIA \cite{belcak2025smalllanguagemodelsfuture} and other researches \cite{fastnav}, we are convinced that SLMs with even smaller params can soon be on par with larger models, given a proper training strategy.

\textbf{SFT and Distillation with Ground Truth Labels.} Standard distillation typically combines the teacher outputs with supervision from true labels, enabling the student model to inherit both the teacher’s knowledge and align with task objectives. However, our Cross Entropy Loss for SFT relies solely on the teacher outputs. This is because our task - path planning in a simulation environment - provides scalar reward signals instead of differentiable ground-truth labels and cannot be directly used in standard loss functions for backpropagation. In our approach, this issue is resolved by leveraging not only SFT but RL, which takes simulation feedback alongside the teacher LLM's outputs to optimize the candidate SLM.

\textbf{Choice of RL Algorithm.} Recent works introduce different alternatives to PPO. For instance, DPO \cite{rafailov2023direct} eliminates the need for an explicit teacher model, making the training process more efficient while often achieving performance comparable to PPO. However, the DPO's objective is fixed by pair-wise preferences and a KL-regularization term, making it difficult to integrate the additional reward components described in Section \ref{ssec: finetuning_pipeline}. In contrast, PPO accepts any scalar reward; we can mix multiple objectives, tune their weights, and monitor their individual effects during training. These reasons make PPO more feasible for our approach. 
\section{Conclusion}
\label{sec:conclusion}
We introduce \textbf{SmallPlan}, a novel LLM-guided distillation framework that leverages SFT and RL to train SLMs for high-level path planning. We discuss our findings on various training strategies and show that the fine-tuned SLMs reach competitive performance compared to superior LLM such as GPT-4o. SmallPlan paves the way for practical deployment of language-driven planning systems on edge devices in real-world robotics.




{
    \small
    \bibliographystyle{IEEEtran}
    \bibliography{main.bib}
}

\end{document}